\documentclass[letterpaper, 10 pt, conference]{ieeeconf}  

\IEEEoverridecommandlockouts                              

\usepackage{graphicx}
\usepackage{cite}
\usepackage{multirow}
\usepackage{array}
\usepackage{float}
\usepackage{amsmath}
\usepackage{amssymb}
\usepackage{hyperref}
\usepackage{xcolor}
\usepackage[labelsep=period, font=footnotesize]{caption}
\usepackage[font=normalsize]{subcaption}
\newcolumntype{P}[1]{>{\centering\arraybackslash}p{#1}}
\newcommand{\RN}[1]{\uppercase\expandafter{\romannumeral #1\relax}}

\begin{document}

\title{\LARGE \bf 
GoonDAE: Denoising-Based Driver Assistance for Off-Road Teleoperation}

\author{Younggeol Cho$^{1}$, Hyeonggeun Yun$^{2}$, Jinwon Lee$^{2}$, Arim Ha$^{2}$, and Jihyeok Yun$^{2}$%
\thanks{Accepted version. This article has been accepted for publication in IEEE Robotics and Automation Letters. DOI: 10.1109/LRA.2023.3250008}
\thanks{\textcopyright 2023 IEEE. Personal use of this material is permitted. Permission from IEEE must be obtained for all other uses, in any current or future media, including reprinting/republishing this material for advertising or promotional purposes, creating new collective works, for resale or redistribution to servers or lists, or reuse of any copyrighted component of this work in other works.}
\thanks{This work was supported by the Agency for Defense Development Grant funded by the Korean Government.
\emph{(Corresponding author: Jihyeok Yun.)}}%
\thanks{$^{1}$Younggeol Cho was with the Agency for Defense Development (ADD), Daejeon 34186, Republic of Korea. He is now with the School of Computing, Korea Advanced Institute of Science and Technology (KAIST), Daejeon 34141, Republic of Korea. (e-mail: rangewing@kaist.ac.kr)}%
\thanks{$^{2}$Hyeonggeun Yun, Jinwon Lee, Arim Ha, and Jihyeok Yun are with the Agency for Defense Development (ADD), Daejeon 34186, Republic of Korea. (e-mail: yhg8423@gmail.com; leejw509@gmail.com; arimida86@gmail.com; jihyeok.yun@gmail.com)}%
}


\maketitle
\thispagestyle{empty}
\pagestyle{empty}

\begin{abstract}
Due to the limitations of autonomous driving technology, teleoperation is used extensively in hazardous environments such as military operations. However, the performance of teleoperated driving is primarily influenced by the driver's skill level. In other words, unskilled drivers need extensive training for teleoperation in harsh and unusual environments, such as off-road. In this letter, we propose GoonDAE, a novel denoising-based real-time driver assistance method that enables stable teleoperated off-road driving. We introduce a denoising autoencoder (DAE) based on a skip-connected long short-term memory (LSTM) to assist the unskilled driver control input through denoising. In this approach, it is assumed that the control input of an unskilled driver is equivalent to that of a skilled driver with noise. We train GoonDAE using the skilled driver control inputs and sensor data collected from our simulated off-road driving environment. Our experiments in the simulated off-road environment show that GoonDAE significantly improves the driving stability of unskilled drivers.
\end{abstract}


\section{INTRODUCTION}
Despite advances in autonomous driving technology, fully autonomous driving is not yet possible due to technological, commercial, and ethical limitations \cite{kettwich2021if, ilkova2017legal, wang2022ethical}. For instance, it is hard to utilize fully autonomous platforms in hazardous environments, such as military operations, where remote platforms must be adjustable and reliable to respond to emergencies in small or large-scale operations \cite{cosenzo2018needs, boukhtouta2004survey} Furthermore, fully autonomous platforms cannot satisfy the military's need for adaptability to urban roads and rugged terrains because most autonomous driving methods presume an urban road with lanes \cite{cosenzo2018needs, naranjo2016autonomous}. Therefore, teleoperation, the remote control of a platform by a human, is preferred for adjustable and reliable operations in various environments \cite{neumeier2019teleoperation, opiyo2021review, lichiardopol2007survey}.

\begin{figure}
\centerline{\includegraphics[width=\columnwidth]{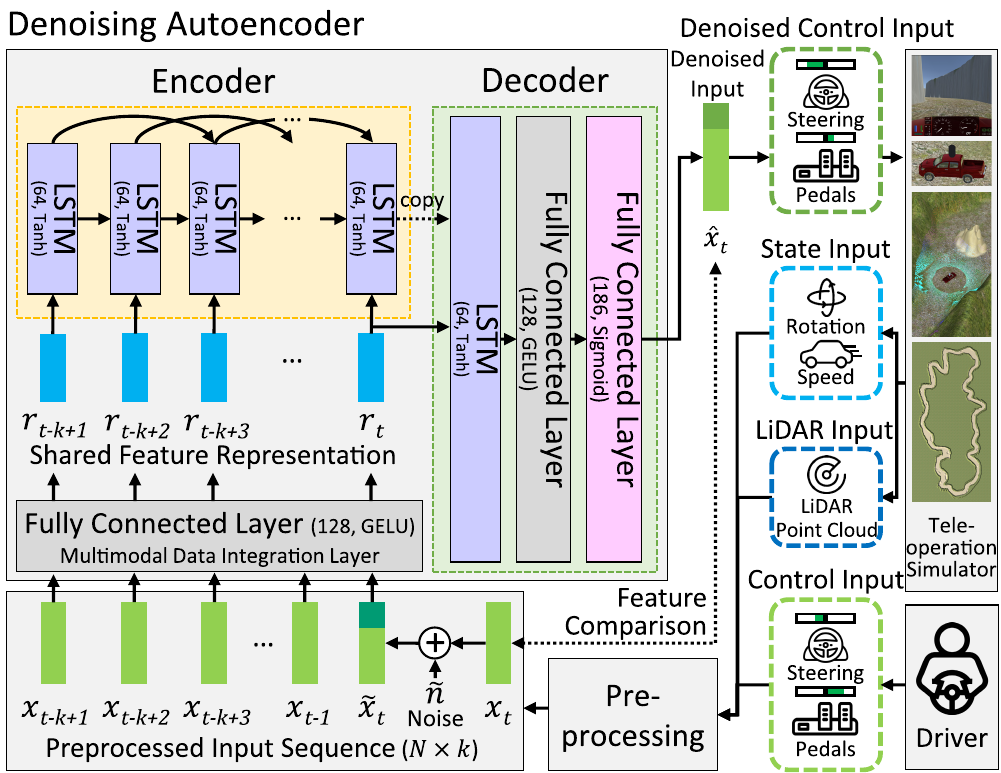}}
\caption{Overview of GoonDAE at training step. GoonDAE denoises driver control input for assistance using a denoising autoencoder based on a skip-connected LSTM. Note that the word goondae stands for the military in Korean.}
\label{fig:model}
\end{figure}

\begin{figure*}[h!]
\centerline{\includegraphics[width=2\columnwidth]{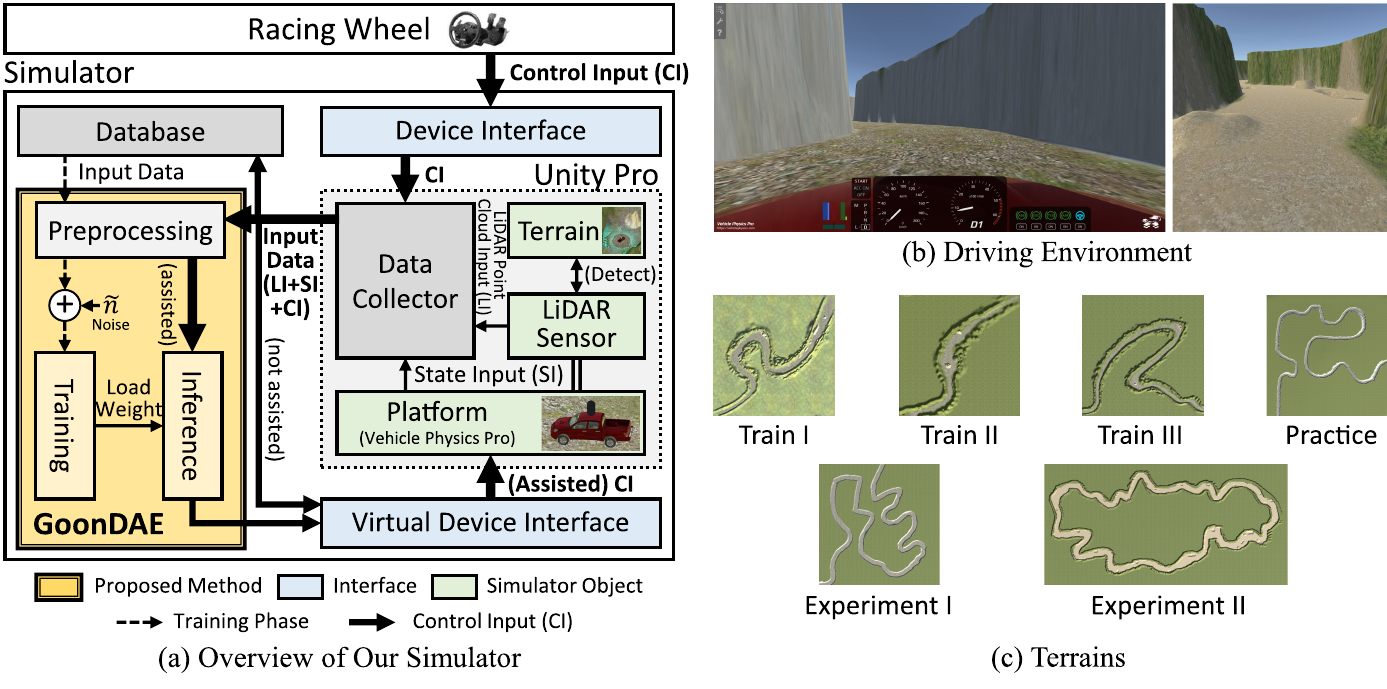}}
\caption{Overview of our simulated teleoperation system.
{(a) Overview of our simulator.} We implemented the simulated environment using an automotive physics engine and a racing wheel. 
{The control input from the racing wheel is delivered to the simulated environment. Control} input (CI), state input (SI), and LiDAR point cloud input (LI) were collected for GoonDAE. 
{The assisted control input from GoonDAE is delivered back to the simulated environment. (b) Driving environment. (c) Off-road terrains: training terrains (Train I, Train II, Train III), a practicing terrain (Practice), and trains for evaluation (Experiment I, Experiment II).}}
\label{fig:simulator}
\end{figure*}

Meanwhile, the user must undertake costly, time-consuming training since teleoperation performance relies on the user's skill level \cite{zheng2020evaluation,chen2007human}. Without training, unskilled drivers {are often unable to complete teleoperated tasks owing} to unstable driving\cite{park2021design, opiyo2021review}. {Unstable driving refers to driving that dramatically varies the platform's state in response to changes in the local environment, such as the emergence of obstacles or a curve\cite{opiyo2021review}.} This issue is exacerbated in challenging and unusual conditions, such as off-road, where teleoperation is predominantly used \cite{diaz2021monitoring, illing2021combining}. To overcome this issue, teleoperation-facilitating driver assistance systems have been developed.

Hacinecipoglu {\em et al.} \cite{hacinecipoglu2013evaluation} introduced haptic feedback based on an obstacle avoidance algorithm to enhance situational awareness during teleoperated driving. Hosseini and Lienkamp \cite{hosseini2016enhancing} proposed a mixed reality interface with a head-mounted display to visualize platform surroundings using sensor data. Although prior literature has shown that teleoperation performance has improved, most studies have only focused on enhancing situational awareness and not assisting driver control input. In other words, it may help skilled drivers by enhancing situational awareness of the drivers; however, {unskilled drivers have difficulty with teleoperation because they cannot effectively utilize the information from the situation.} To sum up, even though approaches centered on situational awareness have been developed, it remains time-consuming to train unskilled drivers.

Shared control is an alternative approach that has been intensively studied for enhancing driving performance. Li {\em et al.} \cite{li2018shared} proposed a shared control driver assistance system based on driving intention recognition and obstacle-avoiding path planning utilizing a model predictive controller. Anderson {\em et al.} \cite{anderson2013intelligent} proposed a semi-autonomous shared control system for obstacle avoidance by restricting driver {control input} depending on the off-road environment. Lei {\em et al.} \cite{lei2022intention} introduced a shared local path planner to facilitate wheelchair motion control based on an intention prediction and user control authorization. {Choe {\em et al.}\cite{choe20141d} proposed a one-dimensional virtual force field (1D-VFF) algorithm that provides obstacle avoidance for real-time local path planning of mobile robots.} However, most studies have only addressed collision avoidance, not driving stability which affects the task accomplishment \cite{park2021design, opiyo2021review}. Consequently, teleoperation typically fails, especially in harsh and unusual environments such as off-road.

In this letter, we propose GoonDAE, a novel denoising-based real-time driver assistance method that enables stable teleoperated off-road driving. Fig. \ref{fig:model}. illustrates an overview of GoonDAE. GoonDAE overcomes the problem of unskilled drivers failing to teleoperate a platform properly in off-road environments in which autonomous driving struggles. Denoising autoencoder (DAE) \cite{vincent2008extracting} is a self-supervised neural network that has been employed extensively for denoising noisy images and audios \cite{deng2014autoencoder, huang2021reduced}  and detecting anomalies in data \cite{park2018multimodal, sakurada2014anomaly}; however, this method has not been applied to assisting {control input}. To apply denoising approach to {control input}, we assume that an unskilled driver {control input} is equivalent to a skilled driver {control input} with noise. GoonDAE constructs an assisted {control input} that emulates a skilled driver from an unskilled driver {control input} by denoising based on the noisy-driving assumption and assists driving with the constructed {control input}. Experimental results revealed that the proposed assistance method considerably enhanced driving performance in terms of driving stability. In particular, the platform was stabilized in terms of speed, lateral position, and collisions with the assistance of GoonDAE in off-road environments where teleoperation and autonomous driving are generally unreliable.

Our contributions are summarized as follows:
\begin{itemize}
    \item We present a real-time, end-to-end driver assistance method, GoonDAE, for teleoperated driving in an off-road environment utilizing self-supervised learning.
    \item We propose a novel driver assistance approach based on a denoising autoencoder and the noisy-driving assumption. To the best of our knowledge, this is the first work introducing denoising to driver assistance.
    \item Our experiments demonstrate quantitatively that the proposed system considerably improves the driving performance of unskilled drivers in terms of driving stability.
\end{itemize}

\section {METHOD}

\subsection{Simulated Teleoperation System}
\label{section:simulator}
{In recent years, the use of vehicle simulators with more precise physics engines has allowed results comparable to experiments conducted in the real world. In order to collect data and evaluate GoonDAE,} we built a simulated teleoperation system, as illustrated in Fig. \ref{fig:simulator}. 
We implemented the 3D simulated teleoperation environment using Unity Pro and Vehicle Physics Pro, 
{a realistic vehicle physics engine, as demonstrated in studies by Park {\em et al.}\cite{park2021toward} and Domberg {\em et al.} \cite{domberg2022deep}. In the studies, the authors used Vehicle Physics Pro to evaluate their suggested approaches for accurate steering control intervention.}
We used the Logitech G29 racing wheel, which includes a steering wheel and pedals, to drive the simulated platform. Functions such as the auto-centered mechanism, traction control system, anti-lock braking system, and electronic stability control, which were supported by the racing wheel and the engine, were applied to simulate a driving experience similar to real-world driving. We set the {maximum rotation degrees} of the steering wheel to 900 degrees. {We set the auto-centering force strength to 40\% and the polling frequency for the racing wheel as 60 Hz. We used the default dynamics parameters setting in Vehicle Physics Pro, as demonstrated in \cite{edy2022vehicle}.} The front view of the platform and vehicle interfaces, such as a speedometer and tachometer, were displayed on a 32-inch monitor to provide a similar experience to a military teleoperation station. We implemented a 64-channel LiDAR sensor imitating the Velodyne HDL-64E and attached it to the platform to collect the surrounding terrain information of the platform. We recorded the rotation and speed of the platform, {control input} from the steering wheel and pedals, and LiDAR point cloud every 100 ms. 

We created {six} canyon terrains, including curves, straight parts, and obstacles, to simulate off-road environments where military teleoperation is typically utilized. {The three Train terrains for collecting teleoperation data for training GoonDAE are based on the real German “KART 2000” stadium track, as utilized by Kim {\em et al.} \cite{kim2022smooth} for learning. The terrain setting was modified to resemble a canyon. The two Experiment terrains for evaluating the performance of GoonDAE were created by adding a high-curvature route, various obstacles, and potholes to the real maneuver test area in our institute \cite{heo2022sim} and the Spanish Army's test track facilities \cite{naranjo2016autonomous} while taking into account their military-purpose uneven ground characteristics.}

\subsection{Denoising-Based Driver Assistance}

We introduced a denoising approach for driver assistance to transform an unskilled driver {control input} into an assisted {control input} that emulates a skilled driver. To achieve this, we assumed that the unskilled driver {control input} was equivalent to the skilled driver {control input} with noise $\tilde{n}$,
\begin{equation}
    CI_{unskilled} = CI_{skilled} + \tilde{n}
\end{equation}
Based on this assumption, we introduced the DAE to provide assistance. We introduced a skip-connected long short-term memory (LSTM) to address real-time sequential inputs based on reports that LSTM-based autoencoders are highly effective in restoring sequential data \cite{srivastava2015unsupervised}. We added skip connections between LSTM layers because they effectively consider the past hidden states of sequential data \cite{kieu2019outlier}.

An overview of GoonDAE is illustrated in Fig. \ref{fig:model}. The DAE in GoonDAE consists of a multimodal data integration layer, an encoder, and a decoder. At each time step $t$, GoonDAE takes the input data {$x_t \in \mathbb{R}^{N}$} with the previous input data ${x}_{t-1}, {x}_{t-2}, ..., {x}_{t-k+1}$ where {the input dimension of GoonDAE $N=186$}; $k$ is the number of time steps, and ${x}_t$ is a concatenation of {control input} vector ${c}_t \in \mathbb{R}^2$, state vector ${s}_t \in \mathbb{R}^4$, and distance vector ${d}_t \in \mathbb{R}^{180}$.
{Control input vector $c_t$ consists of the steering wheel and pedal input. State vector $s_t$ consists of the platform’s yaw, roll, pitch, and speed. Distance vector $d_t$ consists of the closest obstacles’ distances from the platform at each degree azimuth. By processing control input, state input, and LiDAR point cloud input, respectively, the vectors are generated.}

The multimodal data integration layer, which is a fully connected (FC) layer, integrates the multimodal input data ${x}_t$ consisting of {control input}, platform state, and terrain information extracted from {LiDAR point cloud}. The multimodal integration layer produces a shared feature representation ${r}_t \in \mathbb{R}^{128}$,

\begin{figure}
\centerline{\includegraphics[width=\columnwidth]{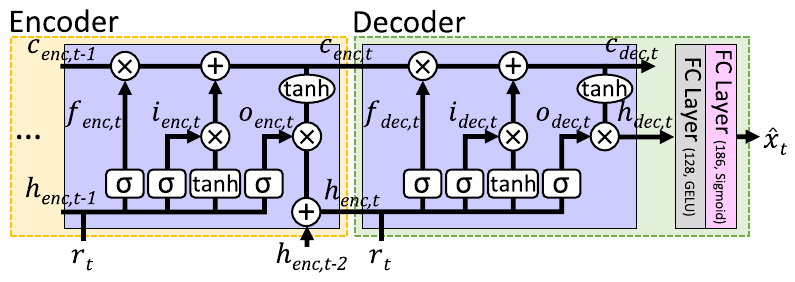}}
\caption{Detailed diagram between the last step of the encoder and the decoder in GoonDAE. The hidden state $h_{enc,t}$ and the cell state $c_{enc,t}$ of the encoder are copied to the decoder to reconstruct the {control input} based on the encoded previous input data.}
\label{fig:encoder_decoder}
\end{figure}

\begin{equation}
    {r}_t = \mathcal{G}({W}_m {x}_t + {b}_m)
\end{equation}
where $\mathcal{G}$ denotes Gaussian error linear unit \cite{hendrycks2016gaussian} function, {${W}_m \in \mathbb{R}^{128 \times N}$} is a weight matrix, and ${b}_m \in \mathbb{R}^{128}$ is a bias.

The encoder, which is an LSTM layer with skip connections, extracts crucial features from the sequential shared feature representations ${r}_{t}, {r}_{t-1}, ..., {r}_{t-k+1}$. The encoder produces the hidden state $h_{enc, t} \in \mathbb{R}^{64}$ and the cell state $c_{enc, t} \in \mathbb{R}^{64}$ at each time step:

\setlength{\arraycolsep}{0.0em}
\begin{eqnarray}
    \label{eq:encoder}
    i_{enc, t} &{}={}& \sigma(W_{enc, i}[h_{enc, t-1}, r_t] + b_{enc, i}) \\
    f_{enc, t} &{}={}& \sigma(W_{enc, f}[h_{enc, t-1}, r_t] + b_{enc, f}) \\
    o_{enc, t} &{}={}& \sigma(W_{enc, o}[h_{enc, t-1}, r_t] + b_{enc, o}) \\
    c_{enc, t} &{}={}& f_{enc, t} \otimes c_{enc, t-1} \\ 
                &&{+}\, i_{enc, t} \otimes \tanh{(W_{enc, c}[h_{enc, t-1}, r_t] + b_{enc, c})} \nonumber \\
    h_{enc, t} &{}={}& o_{enc, t} \otimes \tanh{(c_{enc, t})} + h_{enc, t-2}
\end{eqnarray}
\setlength{\arraycolsep}{5pt}
where $\sigma$ denotes the sigmoid function, $\otimes$ denotes element-wise product, $W_{enc, \bullet}$ and $b_{enc, \bullet}$ are {the weight and bias for the LSTM} $i_{enc, t}$, $f_{enc, t}$, and $o_{enc, t}$ are the input, forget, and output gate signals, respectively. The length of the skip connections was set to 2 to reflect the hidden states before two time steps.

The decoder, which consists of an LSTM layer and two FC layers, denoises the input data to produce denoised {control input}. Because only the current {control input} is required, the data of the latest time step $r_t$ are used as the decoder input. The hidden state $h_{enc, t}$ and the cell state $c_{enc, t}$ are copied from the encoder at the last time step, as illustrated in Fig. \ref{fig:encoder_decoder}. The LSTM layer produces the hidden state $h_{dec, t} \in \mathbb{R}^{64}$ and the cell state $c_{dec, t} \in \mathbb{R}^{64}$:

\setlength{\arraycolsep}{0.0em}
\begin{eqnarray}
    \label{eq:decoder}
    i_{dec, t} &{}={}& \sigma(W_{dec, i}[h_{enc, t}, r_t] + b_{dec, i}) \\
    f_{dec, t} &{}={}& \sigma(W_{dec, f}[h_{enc, t}, r_t] + b_{dec, f}) \\
    o_{dec, t} &{}={}& \sigma(W_{dec, o}[h_{enc, t}, r_t] + b_{dec, o}) \\
    c_{dec, t} &{}={}& f_{dec, t} \otimes c_{enc, t} \\ 
                &&{+}\, i_{dec, t} \otimes \tanh{(W_{dec, c}[h_{enc, t}, r_t] + b_{dec, c})} \nonumber \\
    h_{dec, t} &{}={}& o_{dec, t} \otimes \tanh{(c_{dec, t})}
\end{eqnarray}
\setlength{\arraycolsep}{5pt}
where $W_{dec, \bullet}$ and $b_{dec, \bullet}$ are {the weight and bias for the LSTM} and $i_{dec, t}$, $f_{dec, t}$, and $o_{dec, t}$ are the input, forget, and output gate signals, respectively. The following FC layers finally reconstructs the current input data $\hat{x}_t$,

\begin{equation}
    \hat{x}_t = \sigma(W_{d_2}(\mathcal{G}(W_{d_1}h_{dec, t} + b_{d_1})) + b_{d_2})
\end{equation}
where ${W}_{d_1} \in \mathbb{R}^{128 \times 64}$, {$W_{d_2} \in \mathbb{R}^{N \times 128}$}, ${b}_{d_1} \in \mathbb{R}^{128}$, and {${b}_{d_2} \in \mathbb{R}^{N}$}.

Therefore, the current input data are reconstructed at the output layer of the decoder through the input and LSTM layers. The sigmoid function is used to restore the {control input} within the range 0 to 1. 

\subsubsection{Datasets}

\begin{figure}
\centerline{\includegraphics[width=\columnwidth]{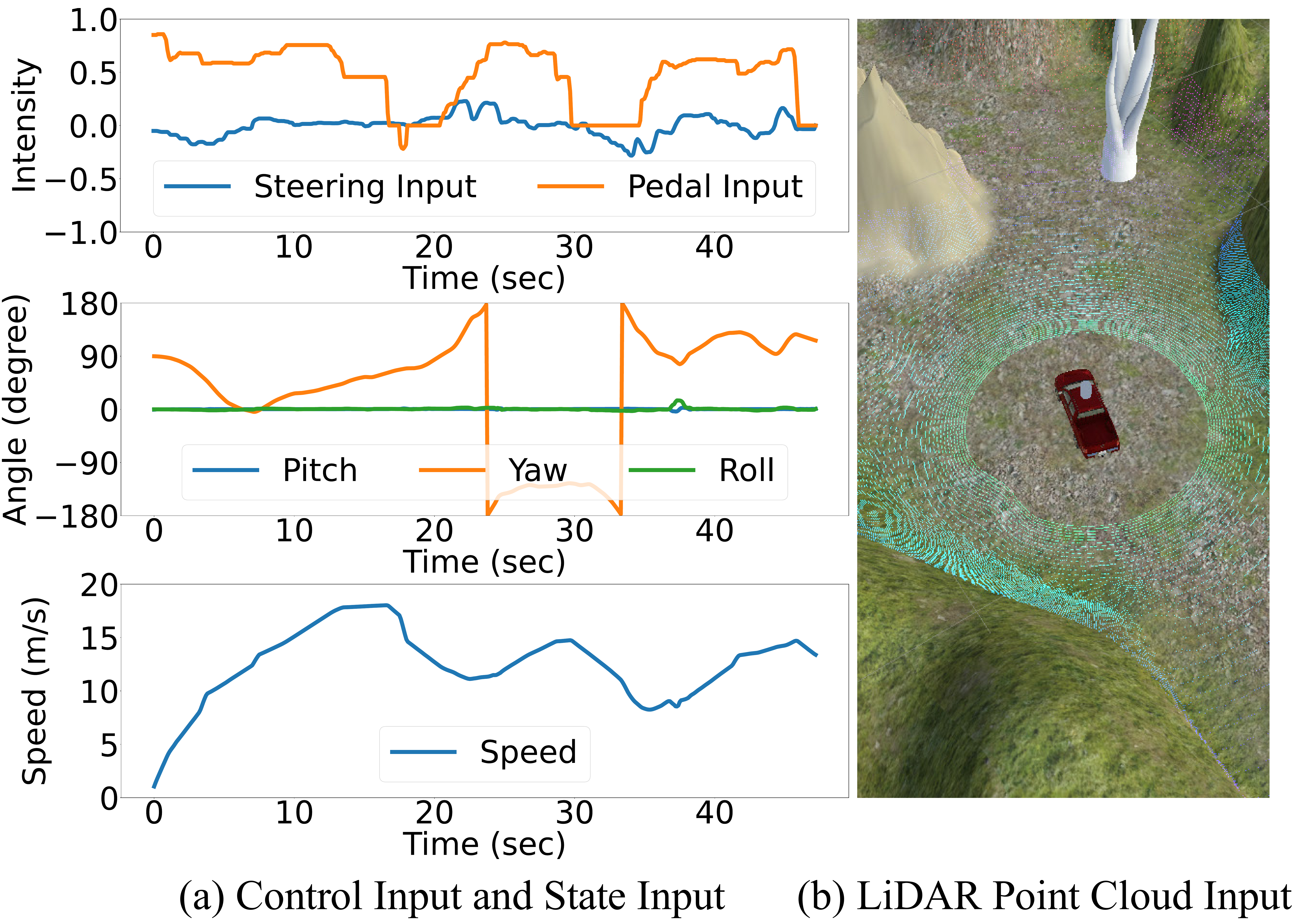}}
\caption{Example of our dataset: {control input, state input, and LiDAR point cloud input.}}
\label{fig:dataset}
\end{figure}

Fig. \ref{fig:dataset} displays example data from our dataset. We collected teleoperation data from {five} skilled drivers by using our simulated teleoperation system. Skilled drivers were requested to teleoperate on the simulated platform. Three canyon terrains were randomly selected for each driving task. The collected data comprised {control input}, {state input}, and {LiDAR point cloud input}, where {control input} is the input from the steering wheel and pedals, {state input} is the speed and rotation of the platform, and {LiDAR point cloud input} is the input from the LiDAR. We combined the inputs from the accelerator and brake pedals by setting the accelerator pedal input value as positive and the brake pedal input value as negative. In conclusion, the pedal input value was between -1 and 1. As a result, in 82 minutes, the driving data of 48965 timesteps were collected and used for training.

\subsubsection{Preprocessing}
We preprocessed the collected data to facilitate training. The preprocessed data consisted of the {control input} vector, state vector, and distance vector. First, the {control input} was normalized to 0--1 to construct the {control input} vector. The {state input} values were normalized to 0--1, assuming a maximum speed of 30 m/s to construct the state vector. The {LiDAR point cloud input} was preprocessed using an obstacle-finding algorithm \cite{choe2016obstacle} after downsampling to 16 channels to reduce the processing time. The distance vector was constructed to a vector of length 180, whose $i$-th entry indicates the distance from the platform center to the nearest obstacle at the corresponding angle, $i$ degree. Only obstacles within the 180-degree range in front of the platform were detected. The distance was normalized to 0--1 by assuming a maximum distance of 50 m.

\subsubsection{Training}
To train GoonDAE, we added noise to the skilled driver {control input} to imitate the unskilled driver {control input}. We trained GoonDAE to produce the assisted {control input} that emulates a skilled driver from unskilled driver {control input} by denoising. We assumed that noise for steering and pedal inputs, $\tilde{n}_{steering}$ and $\tilde{n}_{pedal}$, follows a Gaussian distribution. We added the noises to {control input} vector:

\setlength{\arraycolsep}{0.0em}
\begin{eqnarray}
    \label{eq:noise}
    \tilde{n} &{}={}& (\tilde{n}_{steering}, \tilde{n}_{pedal}) \\
    \tilde{c_t} &{}={}& c_t + \tilde{n}
\end{eqnarray}
\setlength{\arraycolsep}{5pt}
where $\tilde{n}_{steering} \sim N(0, 0.05^2)$ and $\tilde{n}_{pedal} \sim N(0, 0.2^2)$.
We added lower noise to the steering input than the pedal input because the range of the steering wheel was extensive. The noise was added only to the current time step because changing {control input} would change the future state of the platform. 

We used the mean square error function for loss function, however, we only used the error between the {control input} vector $c_t$ and the denoised {control input} vector $\hat{c}_t$: $loss = MSE(c_t, \hat{c}_t)$. We set the total number of time steps $k$ to 10, corresponding to 1 second. We used the PyTorch framework for the experiment. We used 64 batch size and 0.005 initial learning rate, which was decreased to a tenth every 20 epochs, for training. We used Adam as the optimization method and trained the model for 50 epochs on a server with an NVIDIA RTX 3090 GPU.
\begin{table*}[!ht]
  \caption{Comparison of Driving Performances Analyzed by Welch's t-test}
  \label{tab:performance}
  \centering
  \setlength{\tabcolsep}{3pt}
  \setlength{\arrayrulewidth}{0.1mm}
   \begin{tabular*}{\textwidth}{l|@{\extracolsep{\fill}}ccccc|ccccccc}
    \hline
    & \multicolumn{5}{c|}{Experiment \RN{1}}& \multicolumn{7}{c}{Experiment \RN{2}} \\
    & Baseline & GoonDAE & Expert & $\text{p}_{\text{Baseline}}$ & $\text{p}_{\text{Expert}}$ & Baseline & 1D-VFF\cite{choe20141d} & GoonDAE & Expert & $\text{p}_{\text{Baseline}}$ & $\text{p}_{\text{1D-VFF}}$ & $\text{p}_{\text{Expert}}$\\ 
    \hline
        Crash & 1.042 & \textbf{0.417} & 0 & \textbf{0.012} & \textbf{0.000} & 1.318 & 1.182 & \textbf{0.636} & 0 & \textbf{0.039} & \textbf{0.029} & \textbf{0.000}\rule{0pt}{2.2ex}\\
        Frontal Crash & 0.625 & \textbf{0.375} & 0 & 0.210 & \textbf{0.001} & 0.773 & 0.727 & \textbf{0.273} & 0 & \textbf{0.017} & \textbf{0.031} & \textbf{0.011} \\
        Side Crash & 0.417 & \textbf{0.042} & 0 & \textbf{0.006} & 0.328 & 0.545 & 0.455 & \textbf{0.364} & 0 & 0.416 & 0.611 & \textbf{0.008} \\
        SDLP & 1.417 & \textbf{1.268} & 1.220 & \textbf{0.015} & 0.413 & 3.146 & 2.866 & \textbf{2.816} & 2.729 & \textbf{0.001} & 0.464 & 0.271 \\
        SM & 3.672 & \textbf{3.084} & 2.148 & \textbf{0.031} & \textbf{0.000} & 3.652 & 2.990 & \textbf{2.842} & 2.326 & \textbf{0.009} & 0.491 & \textbf{0.040} \\
        TCT & \textbf{166.4} & 169.8 & 144.4 & 0.533 & \textbf{0.000} & \textbf{190.7} & 229.5 & 201.2 & 159.3 & 0.169 & \textbf{0.000} & \textbf{0.000} \\
        ZERO & 0.020 & \textbf{0.018} & 0.014 & 0.180 & \textbf{0.000} & 0.027 & 0.049 & \textbf{0.026} & 0.019 & 0.524 & \textbf{0.000} & \textbf{0.000} \\
    \hline
    \multicolumn{12}{l}{{* $\text{p}_{\text{condition}}$ is a p-value from a pairwise comparison between the GoonDAE and the specified condition.}}\rule{0pt}{2.2ex} \\
  \end{tabular*}
\end{table*}

\subsubsection{Inference}
For driving assistance, the {control input}, {state input}, and {LiDAR point cloud input} were delivered from the simulated system to the model in real-time. The {LiDAR point cloud input} was downsampled to 16 channels before transmission to reduce delays. After preprocessing, the assisted {control input} was obtained using the trained GoonDAE. 
{The assisted control input was generated and delivered to the platform at a rate of 10 Hz, as Naranjo {\em et al.}\cite{naranjo2016autonomous} found that sampling sensor data at a rate of 10 Hz enables real-time obstacle avoidance for a real platform and Kim {\em et al.} \cite{kim2022smooth} adjusted the feedforward steering control to the platform at a rate of 10 Hz. To enable a smooth transition between the 10 Hz assisted control input and the 60 Hz raw control input of the driver, we generated 60 Hz assisted control input. The latest model output and the current 60 Hz raw control input were combined at a ratio of 8:2, following the approach of Choe {\em et al.} \cite{choe20141d}. The combination provides a natural user experience for drivers and helps reduce discomfort from intermittent assistance.}
We conducted this process on a server with an NVIDIA RTX 3090 GPU.

\section{EXPERIMENTS}
To verify the effectiveness of the GoonDAE, we designed and performed {experiments}. 

\subsubsection{Setup}
We conducted two experiments using the simulated teleoperation system described in section \ref{section:simulator}. We used {three canyon terrains for practice, experiment \RN{1}, and experiment \RN{2}, respectively. The distances of the terrains for experiment \RN{1} and \RN{2} were 1.6 km and 1.9 km, respectively, and their} width ranged from 9 to 15 m. A pickup truck was used as the platform to simulate off-road driving. Its width was 1.86 m, and its length was 4.5 m. {In experiment \RN{1},} twelve participants (10 male, 2 female) aged between 20 and 50 years ($\mu$ = 27.75, $\sigma$ = 6.10 years) {participated. In experiment \RN{2}, another twelve participants (10 male, 2 female) aged 20 and 40 years ($\mu$ = 26.50, $\sigma$ = 5.44 years) participated. All participants} participated voluntarily in the experiment. The participants had no prior experience with the simulated system. We received consent for the experiment and made the obtained data anonymous immediately after the experiment.

\subsubsection{Procedure}
We designed a within-subjects study under {three} conditions: Baseline, {1D-VFF,} and {GoonDAE}. In the Baseline condition, participants teleoperated the platform in the simulated teleoperation system without assistance. {In the 1D-VFF and GoonDAE conditions, they performed the task with assistance from the 1D-VFF algorithm and GoonDAE, respectively. Experiment \RN{1} evaluated driving performances under two conditions, Baseline and GoonDAE, while experiment \RN{2} evaluated driving performances under all conditions.} Participants were requested to drive twice for each condition. We shuffled the order of the conditions to remove order effects. The detailed procedure was as follows: (1) \emph{Exercise}–-a participant drove in the terrain for practice until arriving at the end. (2) \emph{Main Experiment}–-a participant drove in the terrain for {each} experiment until arriving at the end, with or without assistance. Every participant completed a total of four drivings {in experiment \RN{1}, and six drivings in experiment \RN{2}. During all of the experiments, participants were instructed to reach the destination as fast and stably as possible, while maintaining a safe distance between the platform and obstacles to prevent collisions.}
{In addition, for comparison purposes, the driving performances of skilled drivers without assistance were assessed (Expert condition).}

\subsubsection{Assessment}
The driving performance {in terms of stability, which refers to the ability to maintain a consistent state when faced with changes in the local environment\cite{opiyo2021review}, } was compared with and without assistance using the following metrics based on the data logged during driving:

\begin{itemize}
    \item Standard deviation of the lateral position (SDLP) \cite{knappe2007driving}: Standard deviation (SD) of the distances from the platform center to the centerline of the lane on the road. It describes how the driver kept the center of the road well, that is, stability of driving. We calculated the distances by determining the difference between the distances to the nearest obstacle on the left and right sides using the LiDAR sensor.
    \item Speed maintenance (SM) \cite{Lewis2011Speed}: SD of the platform speed. It indicates how stable the driving was in terms of acceleration.
    \item Task completion time (TCT): The total driving time from the starting point to the ending point. It indicates how fast the driver can teleoperate the platform.
    \item Number of zero-crossings (ZERO) \cite{knappe2007driving}: The number of changes in the sign (i.e., direction) of the steering wheel angle signal divided by the distance driven. It describes how the driver manipulated the steering wheel left and right, that is, the stability of the steering wheel input.
    \item Number of crashes (Crash): The number of times the platform crashed into an obstacle. Crashes were classified into frontal crashes, which require the platform to reverse to continue after the crash, and side crashes, which do not require reversal.
\end{itemize}

All metrics indicate improved driving performance as they decrease.

\section{RESULTS AND DISCUSSIONS}

\begin{figure*}[ht!]
     \begin{subfigure}[b]{\textwidth}
         \centering
         \includegraphics[width=\columnwidth]{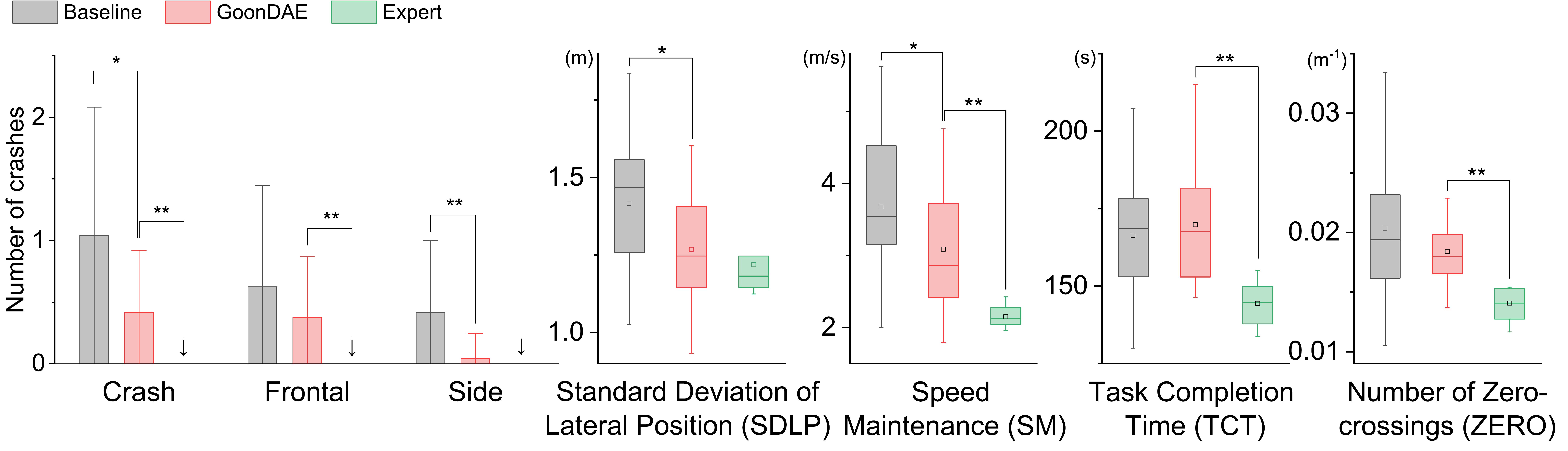}
         \caption{{Experiment \RN{1}}}
         \label{fig:performance_a}
     \end{subfigure}
     \begin{subfigure}[b]{\textwidth}
         \centering
         \includegraphics[width=\columnwidth]{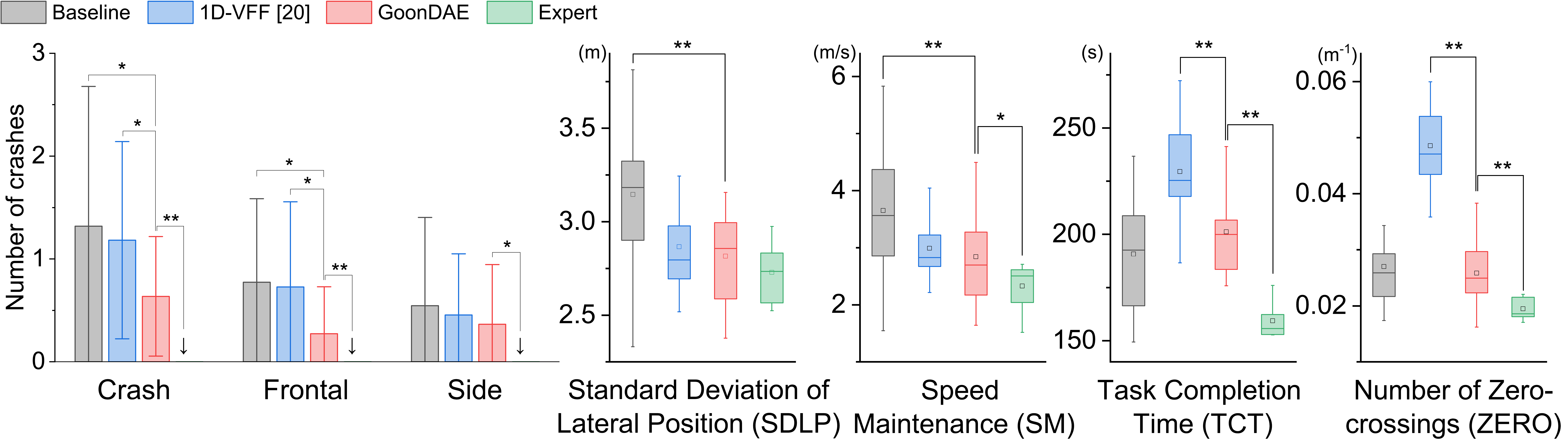}
         \caption{{Experiment \RN{2}}}
         \label{fig:performance_b}
     \end{subfigure}
     
    \caption{Comparison of driving performances. {Welch’s t-test is applied to analyze the statistical significance of GoonDAE-containing pairs.} * denotes p $<$ 0.05 and ** denotes p $<$ 0.01.}
    \label{fig:performance}
\end{figure*}

\begin{figure*}[ht!]
\centerline{\includegraphics[width=2\columnwidth]{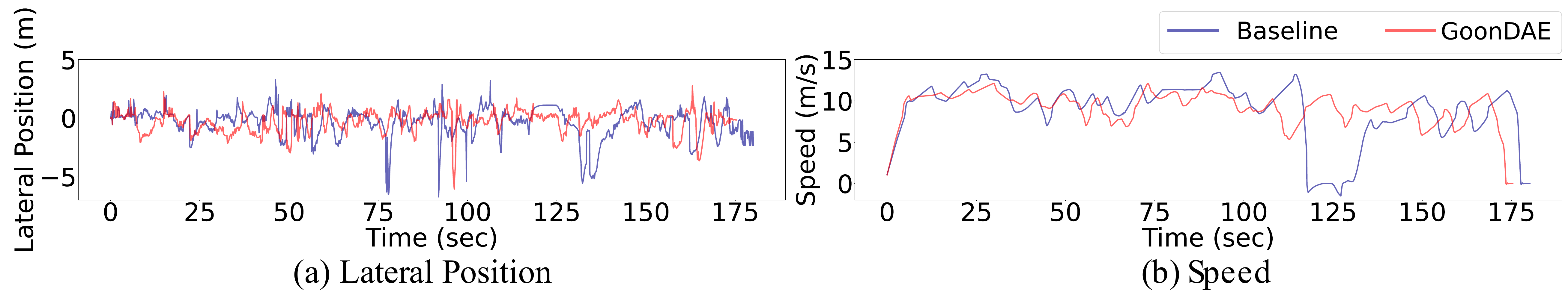}}
\caption{Example of comparison between Baseline and {GoonDAE}. (a) Lateral platform position. (b) Platform {speed.}}
\label{fig:ex_comparison}
\end{figure*}

Fig. \ref{fig:performance} and Table \ref{tab:performance} summarize the experimental results of the driving performances for each case, which were analyzed by Welch's t-test. Fig. \ref{fig:ex_comparison} displays an example of a comparison between Baseline and {GoonDAE. GoonDAE rated Crash, SDLP, and SM considerably lower than the Baseline and the 1D-VFF condition in both experiments. The average ratings for Side Crash and Frontal Crash were lower in GoonDAE in both experiments. Additionally, ZERO was lower in GoonDAE, although} the difference was insignificant. Skilled drivers scored lowest on all criteria except for SDLP, which is comparable to GoonDAE.

The results showed that the proposed assistance method is effective for teleoperation and that our noisy-driving assumption is valid. In particular, the statistically significant differences in SDLP, SM, Crash, Frontal Crash, and Side Crash indicate more stable driving with assistance. A decrease in SDLP revealed that our assistance was effective for steering \cite{knappe2007driving}, and a decrease in SM indicated that it was effective for acceleration and braking. A considerable decrease was observed in {Crash}, implying that it effectively prevents accidents. Contrastly, Crash did not significantly decrease in the 1D-VFF condition. Although no statistically significant differences were observed in TCT and ZERO, the proposed assistance did not have any adverse effects, unlike the 1D-VFF condition, which resulted in lower performance in terms of both TCT and ZERO.

\begin{figure}[t]
\centerline{\includegraphics[width=\columnwidth]{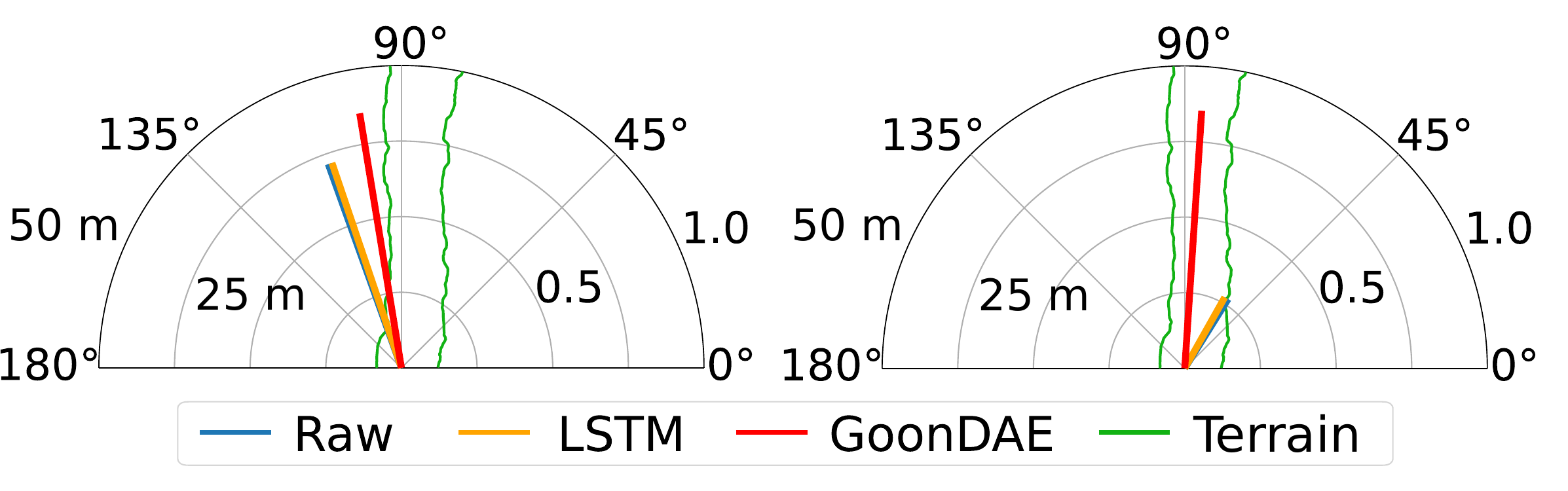}}
\caption{Example of comparison of raw and assisted control inputs. The polar angle refers to the steering input, and the radial distance refers to the pedal input. Green lines denote the shape of the terrain surrounding the platform within 50 m.}
\label{fig:ex_comparison2}
\end{figure}

GoonDAE's performance improvements were not just due to the model imitating skilled drivers. This conclusion was obtained as a result of an experiment in which the same model was trained using the same data and without noise, as typical LSTM. As illustrated in Fig. \ref{fig:ex_comparison2}, the model without noise output equaled the unassisted, raw control input, but GoonDAE output was different. In contrast to LSTM, GoonDAE's output also reflected the raw control input, showing that the denoising enhanced driving performance.

In the user experience survey, 75\% of 12 participants accepted GoonDAE's assistance without difficulty and 41.7\% didn't notice it while performance improved. Only 25\% reported discomfort.

We observed that most frontal crashes occurred in sharp corners after oversteering. Especially in {the GoonDAE condition}, we observed that the steering tended to stay turned even after coming out of a corner. {In experiment \RN{1}, even though the platform collided with far fewer obstacles in front while using GoonDAE, Frontal Crash was not significantly reduced owing to oversteering.
In contrast, in experiment \RN{2}, when the uneven ground characteristic was emphasized rather than sharp corners, Side Crash decreased less compared to Frontal Crash. Despite assistance, the terrain’s harsher uneven ground led the platform’s side to collide with the lateral edge of the obstacles during serial avoidance.}
This phenomenon implied that the proposed system could not provide proper assistance in sharp corners, which appears to be because of a time delay. Therefore, to determine the cause, we measured the end-to-end delay from the racing wheel to the display using the instrument introduced in \cite{cho2020high}. Based on 20 measurements, {the average end-to-end latency was 111 ms, with a maximum of 178 ms. The average teleoperation latency was partly examined as follows: GoonDAE spent 24.3 ms (22 ms for sensor data preprocessing and 2.3 ms for model inference), 47 ms for input handling, rendering, and displaying, and the remaining time for data transmission (1.6 MB/s).}
Furthermore, the {oversteering} was affected by a sampling period of 100 ms.  {In the worst case, the assisted control input may be generated using data sampled 100 ms earlier.}
{Kolar {\em et al.}\cite{kolar2020impact} provided the desired trajectory of linear and angular movement of the platform with control delays. The provided result of Kolar {\em et al.}\cite{kolar2020impact}, about 100 ms of controller-to-actuator delay, supports the practicality of the measured delay. However, previous literature \cite{kolar2020impact, neumeier2019teleoperation, kumcu2017effect} concluded that more than 150 ms of delay decreases teleoperation performance.}
Therefore, a maximum delay of 178 ms can affect driving performance, especially in frontal crashes. Further research is required to overcome the delay problems.

\addtolength{\textheight}{-9cm} 

\section{CONCLUSION}
In this letter, we proposed GoonDAE, a novel denoising-based driver assistance method that improves the driving stability of unskilled drivers during teleoperated off-road driving. GoonDAE produces an assisted {control input} that emulates a skilled drier from an unskilled driver {control input} via denoising, based on the assumption that an unskilled driver {control input} is equivalent to a skilled driver {control input} with noise. Experiment results showed that the suggested approach could help unskilled drivers boost driving stability and prevent crashes. In like manner, the suggested approach could reduce the training time of unskilled drivers by improving their driving stability. In the future, we will consider {the impact of teleoperation latency and drivers' cognitive workload on} the performance of the proposed assistance. {Additionally, we plan to verify the effectiveness of the assistance in a real-world environment using wheeled and tracked vehicles. The assistance will be evaluated on various challenging off-road terrains, including mountains, forests, and plains, which may feature low friction, bumpy roads, and various obstacles, while taking into account the impact of teleoperation latency and the dynamics aspects.}










\bibliographystyle{IEEEtran}
\bibliography{IEEEabrv,main}

\end{document}